\documentclass[runningheads]{llncs}

\usepackage{amsmath}
\usepackage{graphicx}
\usepackage{subcaption}
\usepackage{pgfplots}
\pgfplotsset{compat=newest}
\usepackage{cite}
\usepackage{hyperref}

\usepackage{comment}

\begin{document}

\title{Training Convolutional Neural Networks With Hebbian Principal Component Analysis}

\titlerunning{Hebbian PCA}

\author{Gabriele Lagani\inst{1} \and
Giuseppe Amato\inst{2} \and
Fabrizio Falchi\inst{2} \and Claudio Gennaro \inst{2}}

\authorrunning{G. Lagani et al.}

\institute{
    University of Pisa \\
    \email{gabriele.lagani@phd.unipi.it} \and
    ISTI - CNR, Pisa \\
    \email{giuseppe.amato, fabrizio.falchi, claudio.gennaro @isti.cnr.it}
}

\maketitle

\begin{abstract}
Recent work has shown that biologically plausible Hebbian learning can be integrated with backpropagation learning (backprop), when training deep convolutional neural networks. In particular, it has been shown that Hebbian learning can be used for training the lower or the higher layers of a neural network. 
For instance, Hebbian learning is effective for re-training the higher layers of a pre-trained deep neural network, achieving comparable accuracy w.r.t. SGD, while requiring fewer training epochs, suggesting potential applications for transfer learning.

In this paper we build on these results and we further improve Hebbian learning in these settings, by using a nonlinear Hebbian Principal Component Analysis (HPCA) learning rule, in place of the Hebbian Winner Takes All (HWTA) strategy used in previous work. We test this approach in the context of computer vision. In particular, the HPCA rule is used to train Convolutional Neural Networks in order to extract relevant features from the CIFAR-10 image dataset.

The HPCA variant that we explore further improves the previous results, motivating further interest towards biologically plausible learning algorithms.

\end{abstract}

\section{Introduction}

The error backpropagation algorithm (\textit{backprop}) has been used with great success for training neural networks (e.g. \cite{he, silver}) on a variety of learning tasks, including computer vision. However, Neuroscientists doubt that it is biologically plausible and that it models the real learning processes of the brain \cite{oreilly}. A possible biologically plausible learning mechanisms could be based on the so-called \textit{Hebbian} principle: ``Neurons that fire together wire together". Starting from this simple principle, it is possible to formulate different variants of the Hebbian learning rule which offer interesting opportunities in computer science and artificial intelligence. For example, Hebbian learning with Winner-Takes-All (HWTA) competition \cite{grossberg} allows a group of neurons to learn to perform clustering on a set of data. Another interesting variant is Sanger's rule \cite{sanger}, which allows to perform Principal Component Analysis (PCA) on the data in an online fashion. In essence, Hebbian algorithms can be employed to extract features of interest from data and provide a biologically plausible, efficient and online solution for unsupervised learning tasks.

In the context of Convolutional Neural Networks (CNNs) applied to computer vision, the various layers of the network act as feature extractors, with lower layers extracting low-level features and next layers extracting progressively higher-level features. Hebbian learning algorithms could represent a possible option for training such networks.
Previous works \cite{wadhwa2016a, wadhwa2016b, bahroun} already showed that Hebbian learning variants are suitable for training relatively shallow networks (with two or three layers), which are appealing for applications on constrained devices. In \cite{hebbian_meets_deep, thesis}, it was also shown that HWTA competition was effective to re-train higher layers of a pre-trained network, achieving results comparable with backprop, but requiring fewer training epochs, thus suggesting potential applications in the context of transfer learning. 

In this work, we leverage on these results and we apply a nonlinear Hebbian Principal Component Analysis (HPCA) learning rule \cite{sanger, karhunen} to train CNNs for computer vision tasks.
Specifically, a six layer network is trained using HPCA on the CIFAR-10 \cite{cifar} dataset. We evaluate the quality of the features extracted from each layer on the image classification task by feeding these features to linear classifiers and evaluating the resulting accuracy. The results obtained by the HPCA variant are compared to those obtained by the corresponding network trained with backprop and to HWTA. In order to explore the possibility of integrating HPCA with bakcprop, we also consider hybrid network architectures, in which some layers are trained with backprop, and others with the Hebbian approach.

Results in this paper confirm previous findings showing that Hebbian learning can be integrated with backprop, providing comparable accuracy when used to train lower or higher network layers, while requiring fewer training epochs. Furthermore, we show that the HPCA variant provides further improvements over the results previously obtained with HWTA. 
On the other hand, the current limitation of the approach is that it is not suitable to train intermediate network layers, for which a significant accuracy drop is observed w.r.t. backprop. However, our results suggest that Hebbian approaches are worth being further explored, and we hope that our work can motivate further interest in this direction.

Biologically inspired learning approaches are attractive, in that they could give insights on peculiar features of human intelligence, such as generalization capabilities, learning speed, energy efficiency, robustness to adversarial examples.
Moreover, Hebbian learning algorithms can be easily  adapted  to  Spiking  Neural  Networks  (SNNs)  \cite{gerstner}. These are  neural  network  models  which mimic biological networks  even  more  closely.  Communication  among  neurons is achieved not by means of continuous signals, but by means of  short  pulses. This communication  paradigm  allows  the networks  to  implement  complex  cognitive  functions  with  very limited energy. This makes SNNs suitable for energy  efficient  implementations on neuromorphic hardware \cite{furber2007}, thus enabling intelligent applications also on constrained devices. 
Finally, the Hebbian learning rule is local, which means that every layer of neurons performs updates  independently of the successive layers, encouraging the possibility of modular deep learning and parallel training.

The main contributions of this paper can be summarized as follows:
\begin{itemize}
    \item A nonlinear Hebbian PCA (HPCA) learning rule variant is used to train neural network layers to extract features from an image dataset, which are then used for the task of image classification;
    \item The learning rule is properly integrated with convolutional layers (Convolutional Hebbian PCA);
    \item The results are compared with those obtained by an equivalent network trained with backprop and Hebbian with Winner Takes All (HWTA) and the potentials and limitations of the approach are highlighted;
\end{itemize}

The remainder of this paper is structured as follows: 
Section~\ref{sec:rel_work} provides a background on the related literature;
Section~\ref{sec:hebb_pca} presents the Hebbian PCA rule used in our experiments;
Section~\ref{sec:exp_setup} delves into the details of our experimental setup;
In Section~\ref{sec:results}, the results of our simulations are illustrated;
Finally, Section~\ref{sec:conclusions} presents our conclusions and outlines possible future developments.

\section{Background and related work} \label{sec:rel_work}

Consider a single neuron with weight vector $w$ and input $x$. Call $y = w^T \, x$ the neuron output. The Hebbian learning rule, in its most basic form, can be expressed mathematically as:
\begin{equation}
    \Delta w = \eta \, y \, x
\end{equation}
where $\eta$ is the learning rate. Basically, this rule states that the weight on a given synapse is reinforced when the input on that synapse and the output of the neuron are simultaneously high. Therefore, connections between neurons whose activations are correlated are reinforced.

\begin{figure}
    \centering
    \subfloat[Update step]{
        \includegraphics[width=0.4\textwidth]{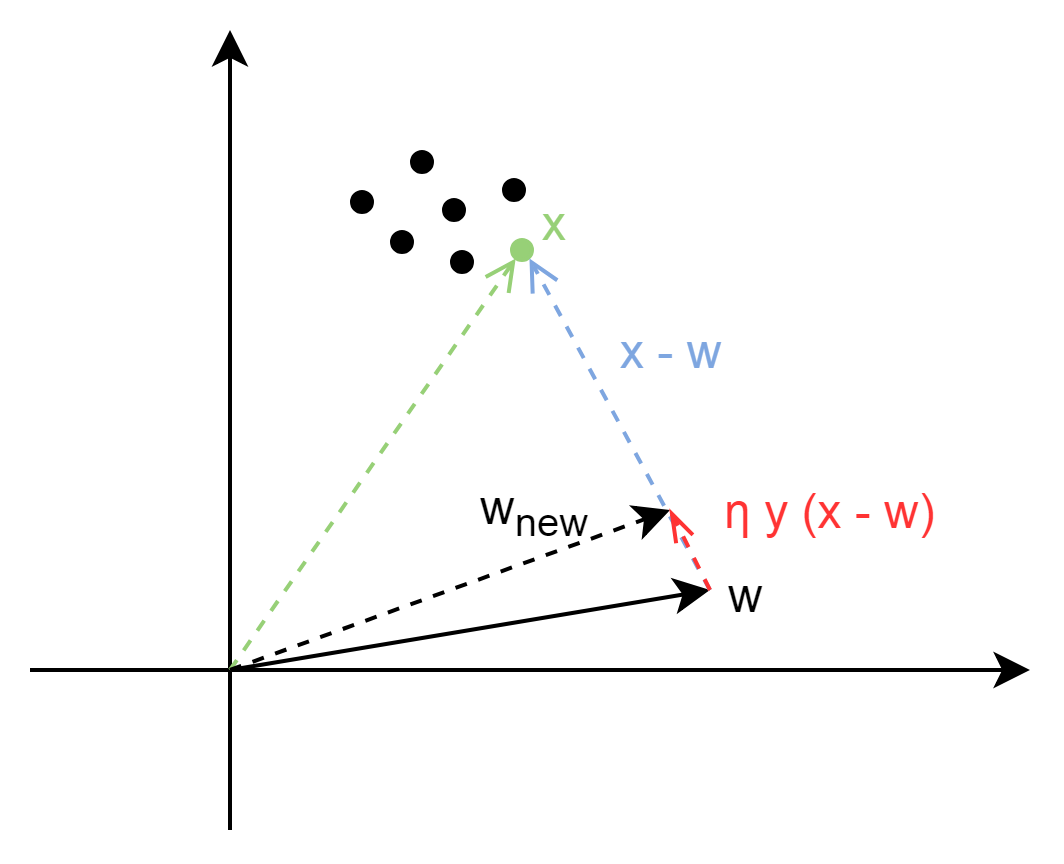}
    }
    ~
    \subfloat[Final position after convergence]{
        \includegraphics[width=0.4\textwidth]{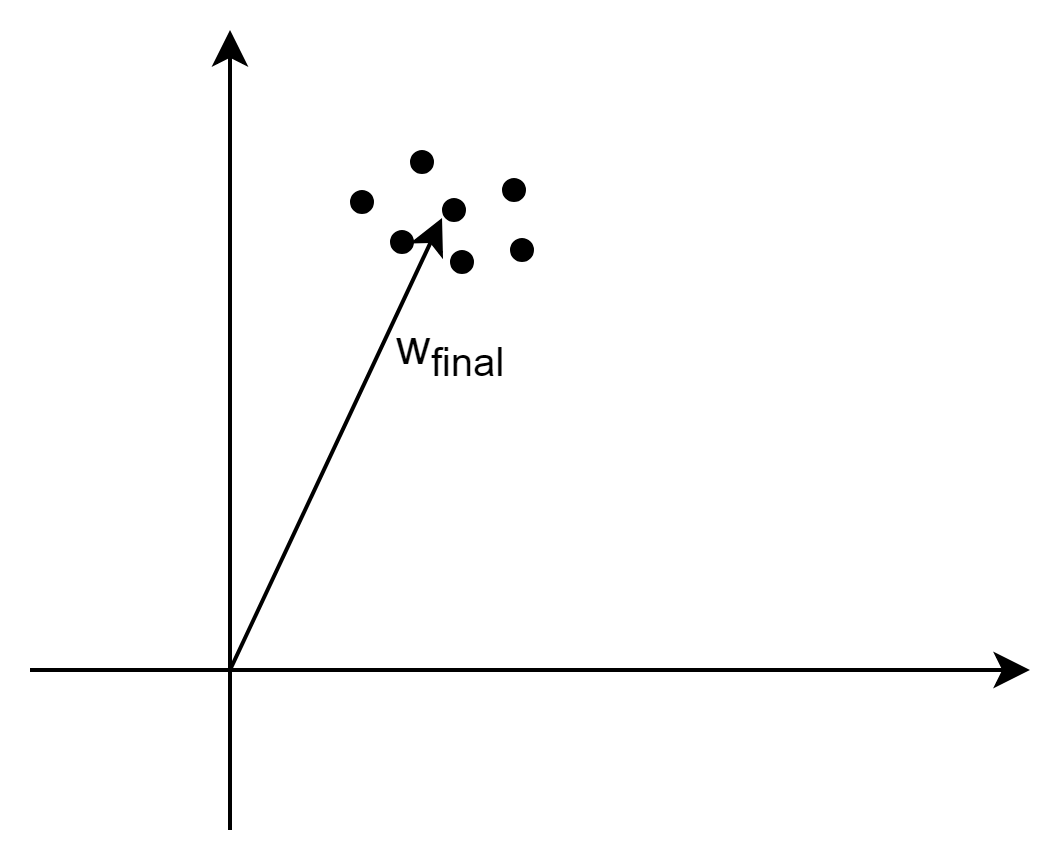}
    }
    \caption{Hebbian updates with weight decay.}
    \label{fig:hebb_update_wd}
\end{figure}

To prevent weights from growing unbounded, a weight decay term is generally added. In the context of competitive learning \cite{grossberg, rumelhart, kohonen1982}, this is obtained as follows:
\begin{equation}
    \Delta w = \eta \, y \, x - \eta \, y \, w = \eta \, y \, (x - w)
\end{equation}
This rule has an intuitive interpretation: when an input vector is presented to the neuron, its vector of weights is updated in order to move it closer to the input, so that the neuron will respond more strongly when a similar input is presented. When several similar inputs are presented to the neuron, the weight vector converges to the center of the cluster formed by these inputs (Fig.~\ref{fig:hebb_update_wd}).

\begin{figure}
    \centering
    \subfloat[Update step]{
        \includegraphics[width=0.5\textwidth]{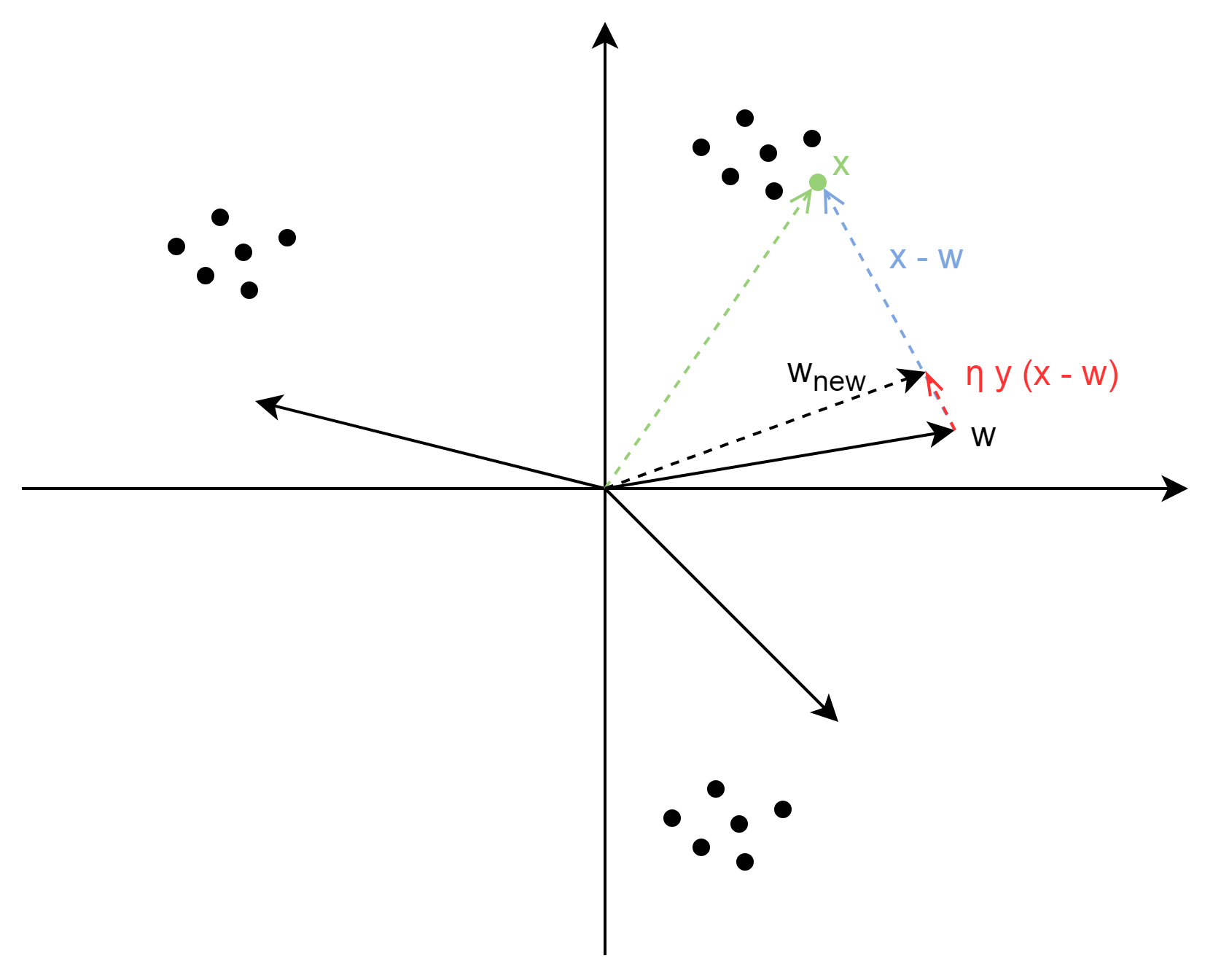}
    }
    ~
    \subfloat[Final position after convergence]{
        \includegraphics[width=0.5\textwidth]{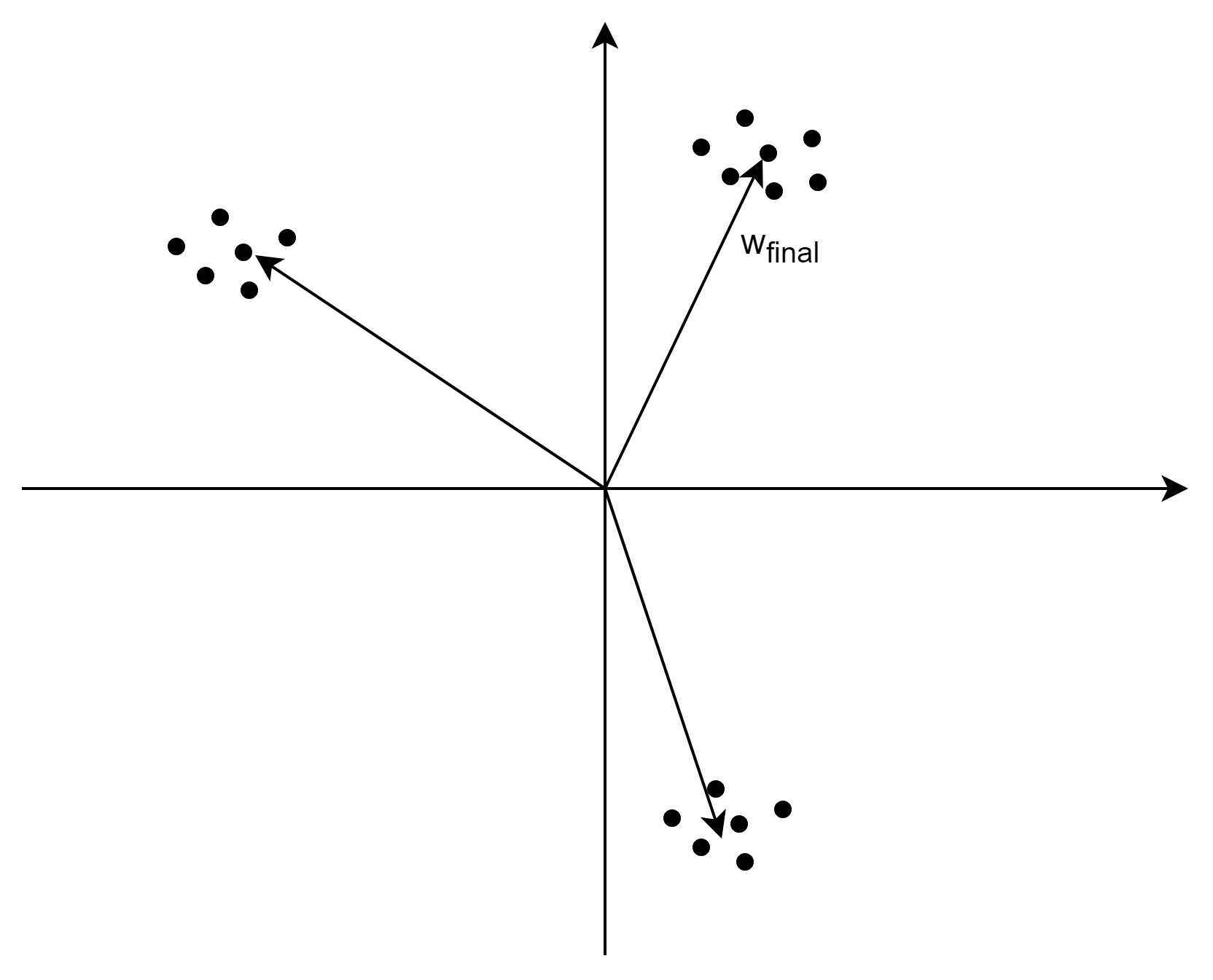}
    }
    \caption{Hebbian updates with Winner-Takes All competition.}
    \label{fig:hebb_update_wta}
\end{figure}

When multiple neurons are involved in a complex network, the Winner-Takes-All (WTA)\cite{grossberg, rumelhart} strategy can be adopted to force different neurons to learn different patterns, corresponding to different clusters of inputs. When an input is presented to a WTA layer, the neuron whose weight vector is closest to the current input is elected as winner. Only the winner is allowed to perform a weight update, thus moving its weight vector closer to the current input (Fig.~\ref{fig:hebb_update_wta}). If a similar input will be presented again in the future, the same neuron will be more likely to win again. This strategy allows a group of neurons to perform clustering on a set of data points (Fig.~\ref{fig:hebb_update_wta}).

In recent works \cite{wadhwa2016a, wadhwa2016b}, WTA and the variant k-WTA (in which the k neurons with highest activations are elected as winners) were applied in the context of computer vision to train a three layer CNN to extract features from images, in order to perform classification. Similar paradigms were also studied in the context of SNNs \cite{ferre, diehl}. These works showed that the approach is suitable to train relatively shallow networks (e.g. with two or three layers), achieving accuracy around 65-70\% on CIFAR-10 \cite{cifar} and from 95\% up to 98-99\% on MNIST \cite{mnist}, which is comparable to backpropagation-based approaches on networks of the same depth.

In \cite{hebbian_meets_deep, thesis}, the authors went further by applying Hebbian-WTA learning to CNNs with up to six layers, comparing the results with those obtained by training the same network with backprop. The WTA approach, as it is, is unsupervised, but a supervised Hebbian learning variant was also proposed in order to train the final classification layer. The results confirmed that the approach was effective for training shallow networks. It was also found that the approach was effective for re-training the higher layers (including the final classifier) of a pre-trained network. In addition, the algorithm required much fewer epochs than backprop to converge.

\section{Hebbian PCA} \label{sec:hebb_pca}
According to the definition given in Section \ref{sec:rel_work}, WTA enforces a kind of \textit{quantized} information encoding in layers of neural network. Only one neuron activates to encode the presence of a given pattern in the input. On the other hand, neural networks trained with back propagation exhibit a \textit{distributed} representation, where multiple neurons activate combinatorially to encode different properties of the input, resulting in an improved coding power. The importance of distributed representations was also highlighted in \cite{foldiak, olshausen}.

A more distributed coding scheme could be obtained by having neurons extract principal components from data, which can be achieved with Hebbian-type learning rules \cite{sanger, becker}. In order to perform Hebbian PCA, a set of weight vectors has to be determined, for the various neurons, that minimize the \textit{representation error}, defined as:
\begin{equation} \label{eq:repr_err}
    L(w_i)  = E[(x - \sum_{j=1}^i y_j \, w_j)^2]
\end{equation}
where the subscript $i$ refers to the $i^{th}$ neuron in a given layer and $E[\cdot]$ is the mean value operator.
It can be pointed out that, in the case of linear neurons and zero centered data, this reduces to the classical PCA objective of maximizing the output variance, with the weight vectors subject to orthonormality constraints \cite{sanger, becker, karhunen}. 
From now on, we assume that the input data are centered around zero. If this is not true, we just need to subtract the average $E[x]$ from the inputs beforehand.

It can be shown that the following learning rule minimizes the objective in eq. \ref{eq:repr_err} \cite{sanger}:
\begin{equation} 
    \Delta w_i = \eta y_i (x - \sum_{j=1}^i y_j w_j)
\end{equation}
In case of nonlinear neurons, a solution to the problem can still be found \cite{karhunen}. Calling $f()$ the neuron activation function, the representation error
\begin{equation}
    L(w_i)  = E[(x - \sum_{j=1}^i f(y_j) \, w_j)^2]
\end{equation}
can be minimized with the following nonlinear version of the Hebbian PCA rule:
\begin{equation} \label{eq:lrn_rule}
    \Delta w_i = \eta f(y_i) (x - \sum_{j=1}^i f(y_j) w_j)
\end{equation}

Several variants of the Hebbian PCA approach were explored in literature for the linear case \cite{sanger, becker, pehlevan_subspace, pehlevan_pca}, and applied in the context of computer vision \cite{bahroun}, but only for shallow networks. In our experiments, we applied the nonlinear version of the Hebbian PCA rule also on deeper networks, as explained in the following sections.

\section{Experimental Setup} \label{sec:exp_setup}

In the following, we describe the details of our experiments and comparisons, discussing the network architectures and the training procedure\footnote{The code to reproduce the experiments described in this paper is available at \url{https://github.com/GabrieleLagani/HebbianPCA}.}.

\subsection{Network architecture and learning}

\begin{figure}
\centering
\includegraphics[width=0.8\textwidth]{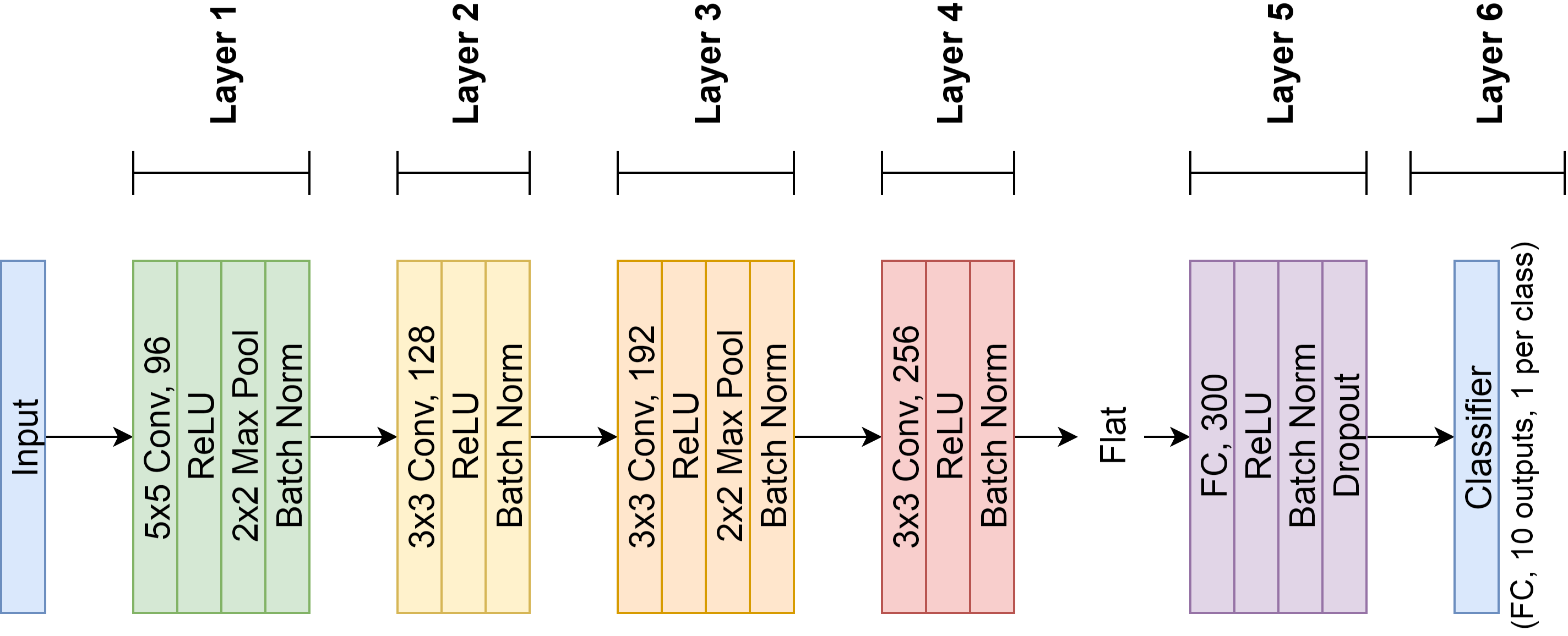}
\caption{The neural network used for the experiments.}
\label{fig:network}
\end{figure}

The core part of our experiments consisted in training the deep layers of a neural network on the CIFAR-10 dataset \cite{cifar}. We used a neural network consisting of six layers: five deep layers plus a final linear classifier. The various layers were interleaved with other processing stages (such as ReLU nonlinearities, max pooling, etc.), as shown in Fig.~\ref{fig:network}. The architecture was the same used in \cite{hebbian_meets_deep, thesis} \footnote{The network achieves 84.95\% accuracy on CIFAR-10 when trained using backprop without data augmentation, and 91.5\% with data augmentation. However, for simplicity, no data augmentation was used in the rest of the experiments described in this paper.}: it was inspired by AlexNet \cite{krizhevsky}, but one of the fully connected layers was removed and, in general, the number of neurons was decreased, in order to reduce the computational cost of the experiments. However, the architectural hyperparameters described above resulted from a parameter search to maximize the accuracy on CIFAR-10 \cite{cifar} obtained from training the model with backprop.

The network was trained on CIFAR-10 using Stochastic Gradient Descent (SGD) with error backpropagation, and with the HPCA rule in eq. \ref{eq:lrn_rule} (in which the nonlinearity was set to the ReLU function), in order to compare the results.

In order to evaluate the quality of the features extracted from the various layers of the trained models for the image classification task, we placed a linear classifier on top of each already trained layer, and we evaluated the accuracy achieved by classifying the corresponding features. This was done both for the SGD trained network and for the HPCA trained network. The linear classifier was trained with SGD in all cases. Notice that this does not raise biological plausibility issues, because backpropagation is not required when SGD is used to train a single layer. Although the Hebbian approach is unsupervised, it is also possible to apply a supervised variant \cite{hebbian_meets_deep, thesis} for training the linear classifier, but we preferred to use SGD in all cases, in order to make comparisons on equal footings. Indeed, the SGD weight update can be considered as a form of supervised Hebbian update, modulated by a teacher signal.

\subsection{Hybrid network models}

We also implemented hybrid network models, i.e. networks in which some layers were trained with backprop and other layers were trained with HPCA, in order to asses up to which extent backprop layers in our model could be replaced with Hebbian equivalent without excessive impact on the accuracy. The models were constructed by replacing the upper layers of a pre-trained network with new ones, and training from scratch using different learning algorithms. Meanwhile, the lower layers remained frozen, in order to avoid adaptation to the new upper layers. Various configurations of layers were considered.

\subsection{Convolutional HPCA}

In order to be able to use the HPCA rule with CNNs, we had to define a proper way to integrate the HPCA rule with convolutional layers. In particular, neurons at different horizontal and vertical offset of the convolutional layer are constrained to have shared weights.

In order to meet this constraint, the learning rule was adapted as follows: each set of neurons looking at the same portion of the image computed their updates by applying rule in eq. \ref{eq:lrn_rule}, the input $x$ being the patch extracted from the image at the specific horizontal and vertical position. We then averaged the updates over the horizontal and vertical dimensions. The resulting update was applied to the kernel shared by all the neurons at different horizontal and vertical locations. When mini-batches of inputs were used during training, the update averaging was performed also over the mini-batch dimension.

\subsection{Details of training}

We implemented our experiments using PyTorch. Training was performed in 20 epochs (although, for the Hebbian approach, convergence was typically achieved in much fewer epochs) using mini-batches of size 64. 

For SGD training, the initial learning rate was set to $10^{-3}$ and kept constant for the first ten epochs, while it was halved every two epochs for the remaining ten epochs. We also used momentum coefficient $0.9$, Nesterov correction, dropout rate 0.5 and L2 weight decay penalty coefficient $6 \cdot 10^{-2}$. 

In the HPCA training, the learning rate was set to $10^{-3}$. No L2 regularization or dropout was used in this case, since the learning method did not present overfitting issues.

The linear classifiers placed on top of the various network layers were trained with supervision using SGD in the same way as we described above for training the whole network, but the L2 penalty term was reduced to $5 \cdot 10^{-4}$.

All the above mentioned hyperparameters resulted from a parameter search to maximize the accuracy on CIFAR-10 in the respective scenarios.

In all the experiments, we used 40000 CIFAR-10 samples for training, 10000 for validation and 10000 for testing, as this is the standard approach with this dataset. In order to obtain the best possible generalization, \textit{early stopping} was used in each training session, i.e. we chose as final trained model the state of the network at the epoch when the highest validation accuracy was recorded.

\section{Results} \label{sec:results}

In Table \ref{tab:classif_on_layers}, we report the CIFAR-10 test accuracy obtained by classifiers placed on top of the various convolutional layers of the network. We compare the results obtained on the network trained with backprop (BP) and HPCA. We also included the results of the Hebbian-WTA (HWTA) method from \cite{hebbian_meets_deep, thesis}, in order make comparisons. In all the cases, the Hebbian approach required fewer training epochs than backprop to converge (1-2 epochs for Hebbian vs 10-20 for backprop).

\begin{table}[t]
    \caption{CIFAR-10 accuracy on features extracted from convolutional network layers (results within 2\% accuracy from BP or higher, but achieved with fewer training epochs, are highlighted in bold).}
    \begin{center}
        \begin{tabular}{|c|c|c|c|}
            \hline
            \textbf{Layer} & \textbf{BP Acc.(\%)} & \textbf{HPCA Acc.(\%)} & \textbf{HWTA Acc.(\%)} \\
            \hline
            Conv1 & 60.71 & \textbf{63.40} & \textbf{63.92} \\
            \hline
            Conv2 & 66.30 & \textbf{65.42} & 63.81 \\
            \hline
            Conv3 & 72.39 & 65.40 & 58.28 \\
            \hline
            Conv4 & 82.69 & 63.60 & 52.99 \\
            \hline
        \end{tabular}
        \label{tab:classif_on_layers}
    \end{center}
\end{table}

The results show a general improvement of the HPCA approach w.r.t. the HWTA approach. In Table \ref{tab:classif_on_layers}, we can observe that both Hebbian approaches reach comparable performance w.r.t. backprop for the features extracted from the first two layers (but in fewer epochs), suggesting possible applications of Hebbian learning for training relatively shallow networks.

The HWTA approach suffers from a decrease in performance when going further on with the number of layers. With the HPCA approach, this problem seems to alleviate, and the accuracy remains pretty much constant when we move to deeper layers. In particular, the HPCA approach exhibits an increase of almost 11\% points w.r.t. HWTA on the features extracted from the fourth convolutional layer. Still, further research is needed in order to close the gap with backprop also when more layers are added, in order to make the Hebbian approach suitable as a biologically plausible alternative to backprop for training deep networks.

In table \ref{tab:hybrid_nets}, we report the results obtained on the CIFAR-10 test set with hybrid networks. In each row, we reported the results for a network with a different combination of Hebbian and backprop layers (the first row below the header represent the baseline fully trained with backprop). We used the letter "H" to denote layers trained using the Hebbian approach, and the letter "B" for layers trained using backprop. The letter "G" is used for the final classifier (corresponding to the sixth layer) trained with gradient descent. The final classifier (corresponding to the sixth layer) was trained with SGD in all the cases (except the last two rows, see later), in order to make comparisons on equal footings. Notice that, as we already said, this does not raise biological plausibility problems, because backprop is not required on the last layer when SGD training is used. The last two columns show the resulting accuracy obtained with the corresponding combination of layers. 

\begin{table}[t]
    \caption{CIFAR-10 accuracy of hybrid network models (results within 2\% accuracy from BP or higher, but achieved with fewer training epochs, are highlighted in bold).}
    \begin{center}
        \begin{tabular}{|c|c|c|c|c|c|c|c|}
            \hline
            \textbf{L1} & \textbf{L2} & \textbf{L3} & \textbf{L4} & \textbf{L5} & \textbf{L6} & \textbf{HPCA Acc.(\%)} & \textbf{HWTA Acc.(\%)} \\
            \hline
            B & B & B & B & B & G & 84.95 & 84.95 \\
            \hline
            H & B & B & B & B & G & \textbf{83.84} & \textbf{84.93} \\
            \hline
            B & H & B & B & B & G & 81.12 & 80.36 \\
            \hline
            B & B & H & B & B & G & 79.50 & 80.68 \\
            \hline
            B & B & B & H & B & G & 81.53 & 80.92 \\
            \hline
            B & B & B & B & H & G & \textbf{83.90} & \textbf{83.75} \\
            \hline
            H & H & B & B & B & G & 78.60 & 78.61 \\
            \hline
            B & H & H & B & B & G & 75.42 & 72.12 \\
            \hline
            B & B & H & H & B & G & 77.00 & 74.98 \\
            \hline
            B & B & B & H & H & G & 79.17 & 76.86 \\
            \hline
            H & H & H & B & B & G & 72.92 & 67.87 \\
            \hline
            B & H & H & H & B & G & 69.38 & 63.68 \\
            \hline
            B & B & H & H & H & G & 68.44 & 62.43 \\
            \hline
            H & H & H & H & B & G & 66.04 & 57.56 \\
            \hline
            B & H & H & H & H & G & 55.91 & 47.24 \\
            \hline
            H & H & H & H & H & G & 54.71 & 41.78 \\
            \hline
            B & B & B & B & B & H & \textbf{84.88} & \textbf{84.88} \\
            \hline
            B & B & B & B & H & H & \textbf{83.47} & \textbf{83.16} \\
            \hline
        \end{tabular}
        \label{tab:hybrid_nets}
    \end{center}
\end{table}

Table \ref{tab:hybrid_nets} allows us to understand what is the effect of switching a specific layer (or group of layers) in a network from backprop to Hebbian training. The first row represents our baseline for comparison, i.e. the network fully trained with backprop. In the next rows we can observe the results of a network in which a single layer was switched. Both HPCA and HWTA exhibit competitive results with the baseline when they are used to train the first or the fifth network layer. A small, but more significant drop is observed when inner layers are switched from backprop to Hebbian, but the HPCA approach seems to perform generally better than HWTA. In the successive rows, more layers are switched from backprop to Hebbian training, and a higher performance drop is observed, but still, the HPCA approach exhibits a better behavior than HWTA. The most prominent difference appears when we finally replace all the network layers with Hebbian equivalent, in which case the HPCA approach shows an increase of 13\% points over HWTA. 
The last two rows aim to show that it is possible to replace the last two layers (including the final classifier) with new ones, and re-train them with Hebbian approach (in this case, the supervised Hebbian algorithm \cite{hebbian_meets_deep, thesis} is used to train the final classifier), achieving accuracy comparable to backprop, but requiring fewer training epochs (2 vs 10, respectively). This suggests potential applications in the context of transfer learning \cite{yosinski}.

\section{Conclusions and future work}
\label{sec:conclusions}

In summary, our results confirm previous findings, i.e. that the Hebbian approach is suitable for training relatively shallow models (e.g. with just one or two layers) or to re-train the final layers of a pre-trained deep neural network, while requiring fewer training epochs. This suggests potential applications in the context of constrained devices, where shallow networks might be appealing (while deeper models could be prohibitive), or in the context of transfer learning, where an experimenter wants to re-train or fine-tune higher network layers of a pre-trained model on a new task. In addition, our results show a general accuracy improvement up to 13\% of the HPCA rule w.r.t. the HWTA method.
On the other hand, we also showed that HPCA is unable to replace SGD for training also intermediate layers of deep neural networks. Further research is still needed to tackle this limitation.

In future work, further improvements might come from exploring more complex feature extraction strategy, which can also be formulated as Hebbian learning variants, such as Kernel-PCA \cite{scholkopf} and Independent Component Analysis (ICA) \cite{hyvarinen}. 
In addition, it would be interesting to move this work also to the context of Spiking Neural Networks (SNNs), where the Hebbian principle is implemented by the Spike Timing Dependent Plasticity (STDP) learning rule \cite{gerstner}. In this case, it is necessary to map the variants of the Hebbian rule to corresponding STDP variants and test their effectiveness for SNN training. 
Another interesting aspect would be to explore in more depth the convergence and generalization properties of Hebbian learning algorithms when few training samples are available. For example, we suggest to train the network on CIFAR-10 using only 5000 or 500 training samples, instead of the whole training set, and compare the the results with SGD. Further experiments should be performed also on other datasets and tasks, in order to obtain a more complete picture of the potentialities of this approach. 
Finally, an exploration on the behavior of such algorithms w.r.t. adversarial examples also deserves attention.

\bibliographystyle{splncs04}
\bibliography{references.bib}

\end{document}